\documentclass[conference]{IEEEtran}
\IEEEoverridecommandlockouts
\usepackage{cite}
\usepackage{amsmath,amssymb,amsfonts}
\usepackage{algorithmic}
\usepackage{graphicx}
\usepackage{textcomp}
\usepackage{xcolor}
\def\BibTeX{{\rm B\kern-.05em{\sc i\kern-.025em b}\kern-.08em
    T\kern-.1667em\lower.7ex\hbox{E}\kern-.125emX}}

\usepackage{hyperref}
\usepackage{graphicx}
\usepackage{subfigure}
\usepackage{booktabs}
\usepackage{multirow}
\usepackage{amsmath}
\usepackage{amsfonts}
\usepackage{amssymb}
\usepackage{bm}
\usepackage{url}
\usepackage{enumitem}

\usepackage[ruled,vlined,linesnumbered]{algorithm2e}
\let\oldnl\nl%
\newcommand{\nonl}{\renewcommand{\nl}{\let\nl\oldnl}}

\SetCommentSty{mycommfont}
\SetNlSty{emph}{}{}

\definecolor{mydarkblue}{rgb}{0,0.08,0.45}
\hypersetup{
 colorlinks=true,
 allcolors=mydarkblue,
}

\begin{document}

\title{Speeding up Computational Morphogenesis with \\ Online Neural Synthetic Gradients}

\author{
Yuyu Zhang$^{\star\S}$, Heng Chi$^{\dagger\S}$, Binghong Chen$^\star$, 
\\ Tsz Ling Elaine Tang$^\dagger$
Lucia Mirabella$^\dagger$, Le Song$^\star$ and Glaucio H. Paulino$^\star$ \\
$^\star$Georgia Institute of Technology \\
$^\dagger$Siemens Corporate Technology \\
$^\S$Email: \texttt{yuyu@gatech.edu,heng.chi@siemens.com} \\
}

\maketitle

\begin{abstract}
A wide range of modern science and engineering applications are formulated as optimization problems with a system of partial differential equations (PDEs) as constraints. These PDE-constrained optimization problems are typically solved in a standard discretize-then-optimize approach. In many industry applications that require high-resolution solutions, the discretized constraints can easily have millions or even billions of variables, making it very slow for the standard iterative optimizer to solve the exact gradients. In this work, we propose a general framework to speed up PDE-constrained optimization using online neural synthetic gradients (ONSG) with a novel two-scale optimization scheme. We successfully apply our ONSG framework to computational morphogenesis, a representative and challenging class of PDE-constrained optimization problems. Extensive experiments have demonstrated that our method can significantly speed up computational morphogenesis (also known as topology optimization), and meanwhile maintain the quality of final solution compared to the standard optimizer. On a large-scale 3D optimal design problem with around 1,400,000 design variables, our method achieves up to 7.5x speedup while producing optimized designs with comparable objectives.
\end{abstract}


\section{Introduction}

Many science and engineering applications in various domains are addressed by turning them into optimization problems constrained by physical laws in forms of a system of partial differential equations (PDEs). The resulting mathematical problems are commonly known as the PDE-constrained optimization~\cite{hinze2008optimization,biegler2003large}. In practice, PDE-constrained optimization is widely used in optimal design
~\cite{zahr2015progressive,zahr2013construction}
, optimal control
~\cite{hu2012approximation,carnarius2007numerical}
, and inverse estimation
~\cite{van20143d,wang2015breast}
problems.
More concretely, PDE-constrained optimization has been applied in many scientific problems such as optical tomography in biomedical imaging~\cite{abdoulaev2005optical}, radiation treatment planning in radiotherapy~\cite{leugering2014trends}, magnetic drug delivery in nanoscience~\cite{antil2018optimizing} and architected materials design in material science~\cite{osanov2016topology}. It has also been used in a variety of engineering problems, such as flow control in fluid mechanics~\cite{gunzburger2003perspectives}, structure design in aerospace, civil and architectural engineering~\cite{zhu2016topology,stromberg2011application,beghini2014connecting}, and optimal design in automobile industry and robotics \cite{wang2004automobile,hiller2012automatic,liu2017optimal}.

PDE-constrained optimization problems are typically solved in a standard discretize-then-optimize approach, which first discretizes the governing PDE into finite-dimensional residual form, and then optimizes the design variables using iterative gradient-based method. In many applications that require high-resolution solution, such as optimal design, the discretized constraints are high-dimensional, which can easily have millions or even billions of variables. When solving such problems, each iteration can take hours or even days to complete, and solving the entire problem typically takes hundreds of iterations~\cite{aage2017giga}. It is of important practical value to speed up large-scale PDE-constrained optimization.

Recent works~\cite{sosnovik2017neural,yu2018deep,banga20183d,DBLP:journals/corr/abs-1901-07761} have attempted to accelerate the optimal structure design by directly predicting the final design using deep learning models. These iteration-free approaches require to collect plenty of training data beforehand, and each training sample is obtained by completely solving a problem with the standard optimizer. These previous works have severe limitations in four aspects. First, collecting sufficient optimization results as training samples is computationally prohibitive, especially for large-scale 3D design problems which can take days to generate a single training sample. Second, due to the fixed dimensionality of training data, their trained models are not able to be generalized to solve new design problems with different design domains. Third, these approaches are not scalable due to the capacity limit of deep learning models. These works only consider 2D or tiny-scale 3D problems with less than 10,000 design variables. Lastly, these approaches are not accurate enough and the predicted final designs tend to have flaws like hanging or disconnected members. This happens because the above-mentioned approaches fail to embed underlying physical laws during the offline learning. Due to these limitations, a general framework for speeding up PDE-constrained optimization that can be accurate, scalable, efficient and generalizable at the same time is much needed.

To address the above challenges, we propose an online neural synthetic gradients (ONSG) framework. Our method is involved in the optimization process, and employs the optimization history to train a deep learning model in online fashion. The deep learning model can generate synthetic gradients to drive the optimization process. Synthetic gradients are much cheaper to obtain, and can replace exact gradients to skip the expensive solve. During the optimization, ONSG switches between applying exact and synthetic gradients. Since the model is trained online, our method is generalizable and can be directly applied to new problems without any pre-collected training data. Moreover, in ONSG, we also propose a novel two-scale optimization scheme, which decomposes the design variables into local subsets but still incorporates global information. With the two-scale scheme, the ONSG can deal with large-scale optimization problems with millions of design variables. We experiment on challenging 3D optimal design problems with more than one million design variables. Our method shows superior performance in terms of accuracy and scalability, which achieves up to 7.5x speedup over the standard optimizer with comparable or even better final objective achieved.

By designing ONSG, we make three key contributions:
\begin{itemize}
	\item We propose a general online learning framework named ONSG for speeding up PDE-constrained optimization. By interacting with the optimization process, our method can be directly applied to new problems without requiring to pre-collect the training data.
	\item We propose a two-scale optimization scheme to decompose the training data into local subsets and incorporate global information. With this technique, the dimensionality of training data can be constant, no matter how large-scale the optimization problem is.
	\item We successfully apply ONSG to computational morphogenesis, which is a challenging and representative application in large-scale PDE-constrained optimization. Our method achieves significant speedup over the standard optimizer with comparable or even better final objective.
\end{itemize}

\section{Related Work}
Solving large-scale PDE-constrained optimization is a computationally intensive task as the computation of exact gradient at each iteration is computationally expensive and, meanwhile, the complete optimization process typically takes hundreds of iterations. Traditionally, to accelerate such large-scale problems, one has to resort to parallel computing \cite{borrvall2001large,aage2015topology,aage2017giga}, advanced iterative solvers \cite{wang2007large,amir2014multigrid}, or multi-scale and multi-resolution approaches \cite{kim2000multi,nguyen2010computational,liu2018efficient}.

More recently, attempts have been made to apply machine learning to accelerate PDE-constrained optimization problems. The surrogate-assisted optimization approaches \cite{haftka2016parallel}, which use various machine learning models such as Gaussian process \cite{easum2017multi} and deep neural networks \cite{briffoteaux2020parallel} to replace expensive function evaluations, are popular to speed up gradient-free optimization problems. However, those approaches are not suitable for topology optimization problems which typically require gradient-based optimizers \cite{sigmund2011usefulness}. In topology optimization, the dominant idea of the existing attempts roots in an iteration-free and offline approach, that is, using machine learning to replace the optimizer either partially or completely so that, once the models are trained, one could directly employ them to the map initial guesses or partially converged solutions to the final one. This idea has been widely explored in optimal structure design problems. For example, Ulu et al. \cite{ulu2016data} developed a data-driven approach for predicting optimized topologies under various loading cases. In this approach, the optimized topologies obtained under a wide range of loading cases are treated as training samples and the a feed-forward neural network is adopted together with the Principal Component Analysis (PCA) to establish a mapping for predicting the optimized topology under a given loading scenario. Sosnovik and Oseledets \cite{sosnovik2017neural} introduced a deep learning-based framework which uses CNN to predict the final optimal design based on two input images: a partially converged design and its change with respect to the previous step. By doing so, the total number of optimization steps are reduced. Later, Banga et al. \cite{banga20183d} took a similar idea and extended the work of Sosnovik and Oseledets to 3D design problems and to incorporate additional inputs such as external loads and boundary conditions. More recently, Yu et al. \cite{yu2018deep} proposed a two-stage prediction procedure to produce nearly-optimal structural design without iterations. When given a design domain, load and boundary condition, the procedure first uses a trained CNN-based encoder and decoder to predict a low resolution structural design and, then, the predicted low-resolution design is mapped to a high resolution one using a trained Generative Adversarial Network (GAN). All the above-mentioned frameworks are proposed in the context of density-based topology optimization approach. With the same goal, Lei et al. \cite{lei2019machine} developed a machine learning-based framework for the Moving Morphable Component (MMC) approach. The framework first uses the MMC approach to generate a set of training samples with different external loads locations and subsequently apply either Supported Vector Regression (SVR) or K-Nearest Neighbors (KNN) to establish an instantaneous mapping between the design parameters and optimized topology. Although achieving partial successes on small- to medium-scale problems, these iteration-free and offline approaches have severe limitations which prevent their success in large-scale problems: 1) they are not \textit{efficient} due to the requirement of collecting a lot of optimization results as training samples; 2) they are not \textit{generalizable} to solve new problems; 3) they are not \textit{scalable} due to the capacity limit of deep learning models; 4) they are not \textit{accurate} and the predicted solution tend to have flaws.

\section{Methodology}

In this section, we first introduce the background of PDE-constrained optimization problems. We then present the online neural synthetic gradients (ONSG) framework to speed up PDE-constrained optimization, which includes an online gradients learning method and a two-scale optimization scheme. As an algorithmic overview, the procedure of speeding up PDE-constrained optimization with ONSG is summarized in Algorithm~\ref{alg:algorithm}.

\subsection{Problem Background}

The general form of PDE-constrained optimization can be expressed as
\begin{equation} \label{eq:PDECO_general_form}
    \begin{aligned}
        \min_{\bm{z},\bm{u}} \quad & \mathcal{J}(\bm{z},\bm{u})  \\
        \text{subject to} \quad & \bm{r}(\bm{z},\bm{u})=\bm{0},
    \end{aligned}
\end{equation}
where $\bm{z} \in \mathbb{R}^{N_d}$ and $\bm{u} \in \mathbb{R}^{N_s}$ are the design variables and the state variables, respectively; $\mathcal{J}(\cdot)$ is the objective function; $\bm{r}(\bm{z},\bm{u})=\bm{0}$ are the discretized PDE constraints. Due to the infinite-dimensional nature of the governing PDE, the standard discretize-then-optimize approach first discretizes it into finite-dimensional residual form using the finite element method~\cite{strang1973analysis,hughes2012finite}. The discretization results in a system of equations denoted as $\bm{r}(\bm{z},\bm{u})=\bm{0}$, which can be either linear or nonlinear in terms of the state variables $\bm{u}$ depending on the problem.

To solve the optimization problem above, the Nested Analysis and Design (NAND) approach is typically used~\cite{biegler2003large,zahr2015progressive}, which treats the state variables $\bm{u}$ as an implicit function of the design variables $\bm{z}$, namely $\bm{u}(\bm{z})$. Thus the original optimization problem in Eq.~\eqref{eq:PDECO_general_form} is recast as
\begin{equation}\label{eq:PDECO_general_form_NAND}
    \begin{aligned}
        \min_{\bm{z}} \quad & \mathcal{J}(\bm{z},\bm{u}(\bm{z})).  
    \end{aligned}
\end{equation}
The NAND approach requires to solve the discretized PDE constraints  $\bm{r}(\bm{z},\bm{u})=\bm{0}$ for the design variables $\bm{z}$ at every iteration in the optimization process. In many industry-scale applications, high-resolution solutions are required, which means that the state variables $\bm{u}$ are usually of high dimensionality, e.g., in the order of million or even higher. If we denote the design and state variables at optimization iteration $k$ as $\bm{z}^{(k)}$ and $\bm{u}^{(k)}$, respectively, the gradients of the objective function $\mathcal{J}(\cdot)$ at that iteration is given by 

\begin{small}
\begin{align}\label{eq:Objective_gradient_NAND}
  &\bm{g}^{(k)}=\frac{\partial \mathcal{J}}{\partial \bm{z}}\big(\bm{z},\bm{u}(\bm{z})\big) \Bigg|_{\bm{z}=\bm{z}^{(k)}}=\frac{\partial \mathcal{J}}{\partial \bm{z}}(\bm{z},\bm{u})\Bigg|_{\substack{\bm{z}=\bm{z}^{(k)}\\\bm{u}=\bm{u}^{(k)}}} \nonumber \\
  &-\Bigg[\bigg(\frac{\partial \bm{r}}{\partial \bm{u}}(\bm{z},\bm{u})\Bigg|_{\substack{\bm{z}=\bm{z}^{(k)}\\\bm{u}=\bm{u}^{(k)}}}\bigg)^{-\top}\frac{\partial \mathcal{J}}{\partial \bm{u}}(\bm{z},\bm{u})\Bigg|_{\substack{\bm{z}=\bm{z}^{(k)}\\\bm{u}=\bm{u}^{(k)}}}\Bigg]\frac{\partial \bm{r}}{\partial \bm{z}}(\bm{z},\bm{u})\Bigg|_{\substack{\bm{z}=\bm{z}^{(k)}\\\bm{u}=\bm{u}^{(k)}}}.
\end{align}
\end{small}

As shown in the equation above, to compute the gradients $\bm{g}^{(k)}$, one needs to solve $\bm{r}(\bm{z}^{(k)},\bm{u})=\bm{0}$ for $\bm{u}^{(k)}$, and compute $(\partial \bm{r}/\partial \bm{u})^{-\top} (\partial \mathcal{J}/\partial \bm{u})$, both of which are very computationally expensive in large-scale optimization problems.

\begin{algorithm}[t]
\caption{Online Neural Synthetic Gradients}
\label{alg:algorithm}
\DontPrintSemicolon
\SetNoFillComment
\SetKwInOut{Input}{input}\SetKwInOut{Output}{output}
\Input{Initialized design variables $\bm{z}^{(0)}$; original discretized PDE constraints $\bm{r}(\bm{z},\bm{u})=\bm{0}$; coarse-scale discretized PDE constraints $\bm{r}_C(\bm{z}_C,\bm{u}_C)=\bm{0}$; objective function $\mathcal{J}(\bm{z}, \bm{u})$; total number of iterations $K$; iterations applying synthetic gradients $\mathcal{I}_s$; deep learning model $\mathcal{M}$}
\Output{Final solution of design variables $\bm{z}^{(K)}$}
$\mathcal{D}_{train} \gets \text{\O}$ \;
\For {$k=0$ \KwTo $K$}{
    $\bm{z}_C^{(k)} \gets$ coarse-grain design variables $\bm{z}^{(k)}$ \;
    $\bm{u}_C^{(k)} \gets$ solve $\bm{r}_C(\bm{z}_C^{(k)},\bm{u}_C)=\bm{0}$ \;
	\If(\tcc*[f]{apply synthetic gradients})
    {$k \in \mathcal{I}_s$}
	{
		$\widetilde{\bm{g}}^{(k)} \gets$ get synthetic gradients predicted by $\mathcal{M}$ \;
    }
	\Else(\tcc*[f]{apply exact gradients})
	{ 
	    $\bm{u}^{(k)} \gets$ solve $\bm{r}(\bm{z}^{(k)},\bm{u})=\bm{0}$ \;
	    $\bm{g}^{(k)} \gets$ get exact gradients with $\bm{u}^{(k)}$ using~\eqref{eq:Objective_gradient_NAND} \;
		$\mathcal{D}_{train} \gets \mathcal{D}_{train} \cup \{\langle \bm{z}^{(k)}, \bm{u}_C^{(k)}, \bm{g}^{(k)} \rangle \}$ \;
		update model $\mathcal{M}$ with training data $\mathcal{D}_{train}$ \;
	}
	$\bm{z}^{(k+1)} \gets$ update $\bm{z}^{(k)}$ with $\bm{g}^{(k)}$ or $\widetilde{\bm{g}}^{(k)}$ \;
}
\end{algorithm}

\subsection{Online Gradients Learning}

To address all limitations of the iteration-free and offline approaches dominated in the literature, we take a completely different path from previous methods. Instead of collecting data beforehand and training model offline, we employ the optimization history as data, and train the deep learning model online. The deep learning model can quickly generate synthetic gradients on-the-fly, and drive the optimization forward. As illustrated in Algorithm~\ref{alg:algorithm}, ONSG switches between applying exact and synthetic gradients. For iterations using exact gradients, ONSG collects data as training samples and updates the deep learning model. For iterations using synthetic gradients as a shortcut, ONSG skips the expensive solve of exact gradients. There are various ways to design $\mathcal{I}_s$, i.e., which iterations are to apply synthetic gradients. We adopt a relatively simple strategy: use exact gradients for the initial $N_I$ iterations to train a decent deep learning model, and afterwards use exact gradients every $N_F$ iterations. The exact gradients serve as the supervision signal to train the deep learning model. Since the model is trained online, the inaccurate prediction at the beginning can be corrected in later iterations. Also, our framework can be directly plugged into new optimization problems, and has no need of any pre-collected training data. Therefore, ONSG is both \textit{efficient} and \textit{generalizable}. We will explain why ONSG is \textit{scalable} in the next section, and show that ONSG is \textit{accurate} in the section of experiments.

\subsection{Two-scale Optimization} \label{subsec:two_scale}

As discussed before, large-scale PDE-constrained optimization problems can easily have millions or even billions of variables in many industry applications. It is very challenging to build a scalable online deep learning model for such problems. The reasons are three folds: 1) as the number of variables increases, the dimensionality of the input and output grows, so that learning the mapping from input to output becomes much more difficult due to the limited model capacity; 2) such increase of learning difficulty may be alleviated in small-scale problems by providing more supervision, i.e., collecting exact gradients in more iterations for model training. However, for large-scale problems, evaluating exact gradients is very time-consuming, and it can significantly hurt the speedup performance if more supervision is provided; 3) the size of deep learning model grows together with the dimensionality of input and output, which costs more GPU memory for training and predicting. Therefore, the maximum GPU memory available bounds the scale of the optimization problem that can be handled.

To address the challenges above in the single-scale deep learning model, we propose a novel two-scale optimization scheme, which decomposes the training data into local subsets but still incorporates global information. More specifically, we segment the design variables $\bm{z}$ into small subsets by grouping them based on their physical locations in the design space, and each subset serves as one training sample. Based on the segmentation, we also introduce the corresponding coarse-grained design variable vector $\bm{z}_C$ so that each of its component is the averaged value of $\bm{z}$ in each subset. Since we are optimizing all design variables together, it is not feasible to build a deep learning model using only local information within each subset. To incorporate global information, we introduce coarse-scale discretized PDE constraints $\bm{r}_C(\bm{z}_C,\bm{u}_C)=\bm{0}$ and solve it for $\bm{u}_C$, which is used as additional features for each subset of $\bm{z}$.
In this way, we maintain a constant dimensionality of the input and output spaces, thus a constant number of model parameters, no matter how large-scale the optimization problem is. In our experiments, we quantitatively show the scalability of the two-scale optimization scheme.

\begin{figure*}[t]
\centering
\includegraphics[width=\textwidth]{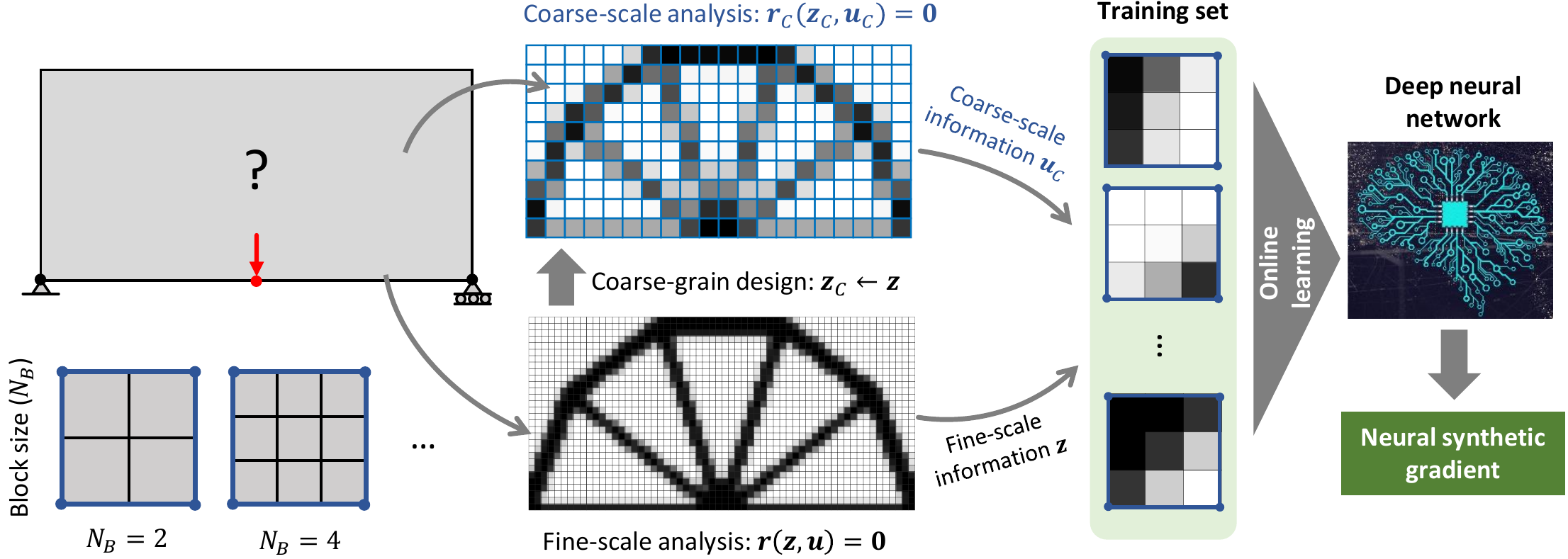}
\caption{Illustration of our two-scale optimization scheme, which groups the design variables into a coarse-scale mesh based on their physical locations in the design space. For the computational morphogenesis problem, we incorporate both global and local information from the coarse-scale and fine-scale mesh to provide input for training the deep learning (DL) model to generate synthetic gradients. As illustrated in the bottom left corner, the block size $N_{B}$ determines the ratio between the side of an element of the fine-scale mesh to the side of an element of the coarse-scale mesh.}
\label{fig:two-scale}
\end{figure*}

\subsection{ONSG for Computational Morphogenesis}

In this section, we showcase the application of ONSG on a challenging and representative class of PDE-constrained optimization problems: the optimal structure design~\cite{bendsoe2013,aage2017giga}. The goal of the optimal structural design is to find the best structure that can sustain a given load. The corresponding PDE-constrained optimization formulation is stated as~\cite{bendsoe2013}:
\begin{equation}\label{eq:min_compliance}
    \begin{aligned}
        \min_{\bm{z}} \quad & \mathcal{J}(\bm{z},\bm{u}(\bm{z})) = \bm{f}^\top \bm{u}(\bm{z}) \\
        \text{subject to} \quad & \mathcal{V}(\bm{z})=\bm{v}^\top\mathbf{P}\bm{z}-V_{\text{max}} \leq 0 \\
        & 0 \leq z_i\leq 1 \quad \forall i \in \{1, \ldots, N_d\},
    \end{aligned}
\end{equation}
where $\bm{v}$ is a constant vector collecting element volumes, $\mathbf{P}$ is the density filter matrix \cite{bendsoe2013} and $V_{\text{max}}$ is the maximum volume imposed on the design. In the about NAND formulation, the implicit function $\bm{u}(\bm{z})$ is defined through the discretized PDE constraints $\bm{r}(\bm{z},\bm{u})=\mathbf{K}(\bm{z})\bm{u}-\bm{f}=\bm{0}$, where $\mathbf{K}$ and $\bm{f}$ are the stiffness matrix and the global force vector, respectively. According to Eq.~\eqref{eq:Objective_gradient_NAND}, the gradients of the objective function $\mathcal{J}(\cdot)$ at iteration $k$ is computed as
\begin{equation}
    g_i^{(k)}=-\big(\bm{u}^{(k)}\big)^{\top}~\frac{\partial \mathbf{K}}{\partial z_i}\big(\bm{z}^{(k)}\big)~\bm{u}^{(k)},
\end{equation}
where $g_i^{(k)}$ is the $i$-th component of $\bm{g}^{(k)}$. We emphasize that, to compute the gradient $\bm{g}^{(k)}$, one needs to first solve $\bm{u}^{(k)}$ through $\mathbf{K}(\bm{z}^{(k)})\bm{u}-\bm{f}=\bm{0}$, which is very computationally expensive for large-scale optimization problems. Iterative solvers such as preconditioned conjugate gradient (PCG) and multi-grid methods are typically used. On the other hand, the gradient of the volume constraint function $\mathcal{V}$ is simply given by $\partial \mathcal{V}/\partial \bm{z} =\textbf{P}^{\top}\bm{v}$, which does not require expensive computation. Once the gradients are obtained, we use the optimality criteria (OC) method~\cite{bendsoe2013}, which the standard optimizer for \eqref{eq:min_compliance}, to update design variables $\bm{z}$. 

To apply the two-scale optimization scheme to computational morphogenesis, we segment the design variables $\bm{z}$ into local subsets (3D blocks) of uniform dimensions $N_B \times N_B \times N_B$, where $N_B$ is the block size, as illustrated in Figure~\ref{fig:two-scale}. The resulting coarse-scale discretized PDE constraints become $\bm{r}_C(\bm{z}_C,\bm{u}_C)=\mathbf{K}_C(\bm{z}_C)\bm{u}_C-\bm{f}_C=\bm{0}$, where $\mathbf{K}_C$ and $\bm{f}_C$ are the coarse-scale stiffness matrix and the force vector, respectively.

\subsection{Model Architecture}

We explored several different architectures for the deep learning model, including fully-connected neural networks, convolutional neural networks and its variants residual networks~\cite{he2016deep}. We finally choose fully-connected DenseNet~\cite{huang2017densely} since it performs the best on various design domains. Our model connects all layers in a fully-connected neural network and preserve the feed-forward nature. To be more specific, each layer obtains additional inputs from all preceding layers and passes on its own output to all subsequent layers. We concatenate the incoming features to serve as input of each layer. Our model is different from DenseNet: we use fully-connected layers while DenseNet uses convolutional layers. The hyperparameters of our model, such as the number of layers, learning rate and hidden layer size, are provided in the experiment section.

\section{Experiments} \label{sec:exp}

In this section, we conduct extensive experiments to demonstrate the effectiveness of ONSG in speeding up large-scale PDE-constrained optimization problems. We first show the scalability of the proposed ONSG framework by increasing the scale of the optimization problem. Then, we study the impact of an important ONSG hyperparameter, the block size $N_B$, which can further speedup the optimization process. We evaluate ONSG by tracking the objective achieved along with the wall-clock time. Also, we plot the final solution for each benchmark problem. For all the experiments, we fairly compare ONSG with the standard optimizer, running with exactly the same computational resources.

\begin{figure}
\centering
	\subfigure[Cantilever]{
	\includegraphics[width=0.16\textwidth]{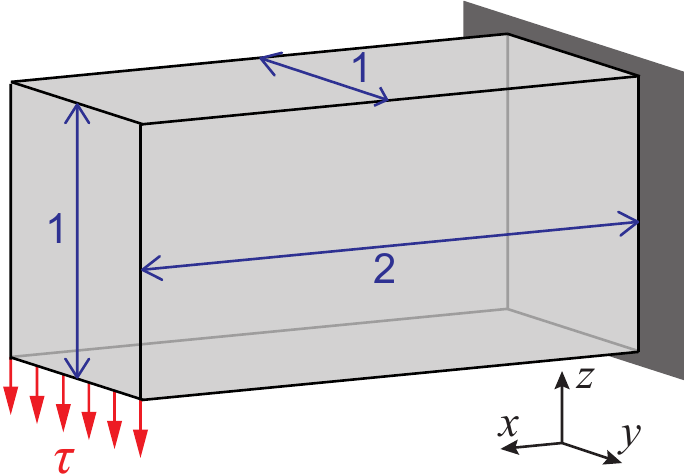}
	\label{fig:cantilever_domain}
	}
	\hfill
	\subfigure[MBB]{
	\includegraphics[width=0.28\textwidth]{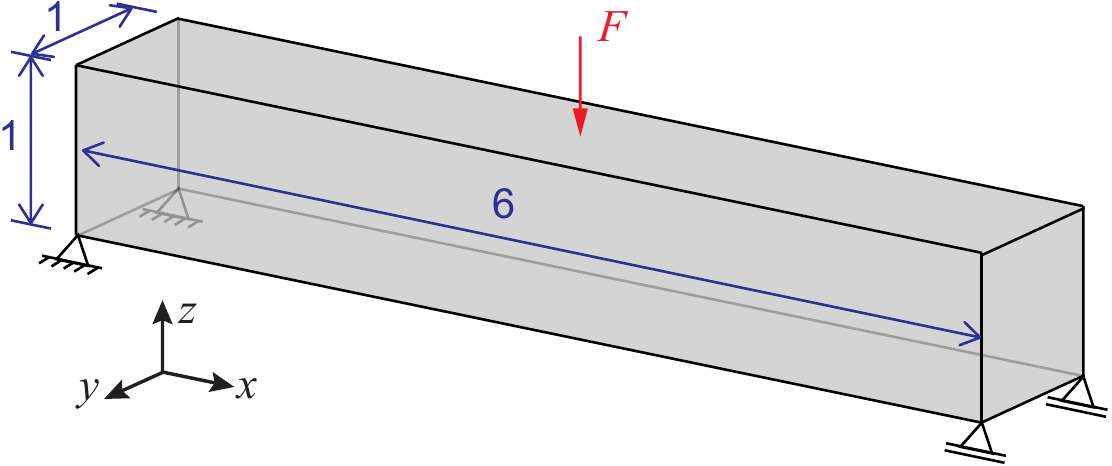}
	\label{fig:MBB_domain}
	}
\caption{Illustration of the design domains of two benchmark problems.}
\label{fig:domain}
\end{figure}

\noindent \textbf{Benchmark problems.} We aim to construct a challenging benchmark, and choose two large-scale PDE-constrained optimization problems in optimal structure design: 1) a cantilever beam design problem, whose dimensions and boundary conditions are illustrated in Figure~\ref{fig:cantilever_domain}. The design domain is fixed on its face $x = 0$ and subjected to a distributed load $\tau=1$ in the negative $z$ direction at the lower edge of the face $x = 2$. We construct three fine-scale meshes (Mesh 1--3) from lowest to highest level of refinement. The volume fraction is $V_{\text{max}} = 12\%$, and the density filter radius is $R = 0.08$; 2) a Messerschmitt–B\"olkow–Blohm (MBB) beam design problem, whose dimensions and boundary conditions are illustrated in Figure~\ref{fig:MBB_domain}. A downward force $F=1$ is applied on the center of the top surface of the design domain. We construct a highly refined mesh (Mesh 4) for this problem. The volume fraction and the density filter radius are set to $V_{\text{max}} = 12\%$ and $R = 0.08$, respectively. Both are large-scale optimal design problems in 3D, and their refined meshes have more than one million design variables. One can refer to Table~\ref{table:stats} for the exact number of design variables (\# DVs) and the dimension of the stiffness matrix $\mathbf{K}$ of each mesh.

\setlength\tabcolsep{4pt}
\begin{table}[t]
\centering
\caption{Statistics of the benchmark problems.}
\label{table:stats}
\begin{tabular}{@{}lcrrcrr@{}}
\toprule
\multirow{3}{*}{\textbf{}} & \multicolumn{3}{c}{\textbf{Cantilever}} & \multicolumn{3}{c}{\textbf{MBB}} \\ \cmidrule(l){2-7} 
 & Mesh 1 & \multicolumn{1}{c}{Mesh 2} & \multicolumn{1}{c|}{Mesh 3} & \multicolumn{3}{c}{Mesh 4} \\ \cmidrule(l){2-7} 
 & \multicolumn{3}{c|}{$N_B=5$} & $N_B=2$ & \multicolumn{1}{c}{$N_B=4$} & \multicolumn{1}{c}{$N_B=8$} \\ \midrule
\# DVs & \multicolumn{1}{r}{86K} & 250K & \multicolumn{1}{r|}{1.5M} & \multicolumn{1}{r}{1.4M} & 1.4M & 1.4M \\
$\textbf{K}$ dim & \multicolumn{1}{r}{276K} & 788K & \multicolumn{1}{r|}{4.5M} & \multicolumn{1}{r}{4.1M} & 4.1M & 4.1M \\
$\textbf{K}_C$ dim & \multicolumn{1}{r}{3K} & 8K & \multicolumn{1}{r|}{40K} & \multicolumn{1}{r}{71K} & 10K & 3K \\ \bottomrule
\end{tabular}
\end{table}

\noindent \textbf{Competitor method.} We compare our ONSG method with the standard optimizer in the literature of large-scale structural design~\cite{aage2017giga,aage2015topology}. Note that the standard optimizer also employs GPUs to solve the discretized PDE constraints, which is the most computationally intensive part in the optimization process. The standard optimizer uses the PCG method with the Jacobi preconditioner~\cite{saad2003iterative}, which is shown to be the state of the art in terms of efficiency~\cite{duarte2015polytop++}. We use the official GPU parallel computing implementation of the PCG method in Matlab.

\noindent \textbf{Experimental settings.} All the experiments are conducted on a GPU-enabled (Nvidia Titan Xp) Linux machine powered by Intel Xeon Silver 4116 processors at 2.10GHz with 256GB RAM. The GPU memory is 12GB. To ensure fair comparison, we allocate exactly the same computational resources (CPU, GPU and memory) for all the experiments. For ONSG experiments, we fix the block size $N_B=5$ on three Cantilever meshes, and have different block sizes $N_B=2,4,8$ on the MBB mesh. For the deep learning model in ONSG, we use a fully-connected neural network with 4 hidden layers. The size of each hidden layer is set to 1000 for the Cantilever experiments, and set to $200,500,1000$ for the MBB experiments with $N_B=2,4,8$ respectively. The learning rate is set to 0.0005 initially, and decays by half every 500 iterations. We train the neural network for 2000 iterations for the Cantilever experiments, and 3000 iterations for the MBB experiments. We use $N_I=10$ initial iterations with exact gradients. We fix the online update interval $N_F=10$ for the Cantilever problems, and set $N_F=50,45,10$ for the MBB experiments with $N_B=2,4,8$ respectively. We set the total number of iterations $K=200$ for all the experiments.

\noindent \textbf{Evaluation metrics.} We use the optimization objective to evaluate the model accuracy. Since we are minimizing the compliance, the objective is the lower the better. To evaluate the model efficiency, we time each experiment and report the wall-clock time in seconds. We also track the intermediate objective during the optimization process, and plot the objective curve along with the wall-clock time. With the plot, we compare the objective achieved by different methods given certain amount of time. Also, we compare the time cost of different methods to achieve a certain objective value.

\subsection{Model Accuracy}

As stated in Eq.~\eqref{eq:min_compliance}, the goal of the design problems is to minimize the compliance of the structure, which is the objective of the optimization process. Thus, the objective is the lower the better. Our method uses the online neural synthetic gradients to replace the standard gradients, which will guide the optimization to a different path. So we first check the accuracy of our method, i.e., how close it is to the standard optimizer in terms of the final objective achieved. We report the experimental results in Table~\ref{table:objective}. For all the cases, ONSG achieves comparable or even better objective, compared to the standard optimizer. Since the optimization problem is highly non-convex, ONSG is able to drive the optimization to a better local minima and beat the standard optimizer to get negative difference in many cases.

\begin{figure}[t]
\centering
\includegraphics[width=0.48\textwidth]{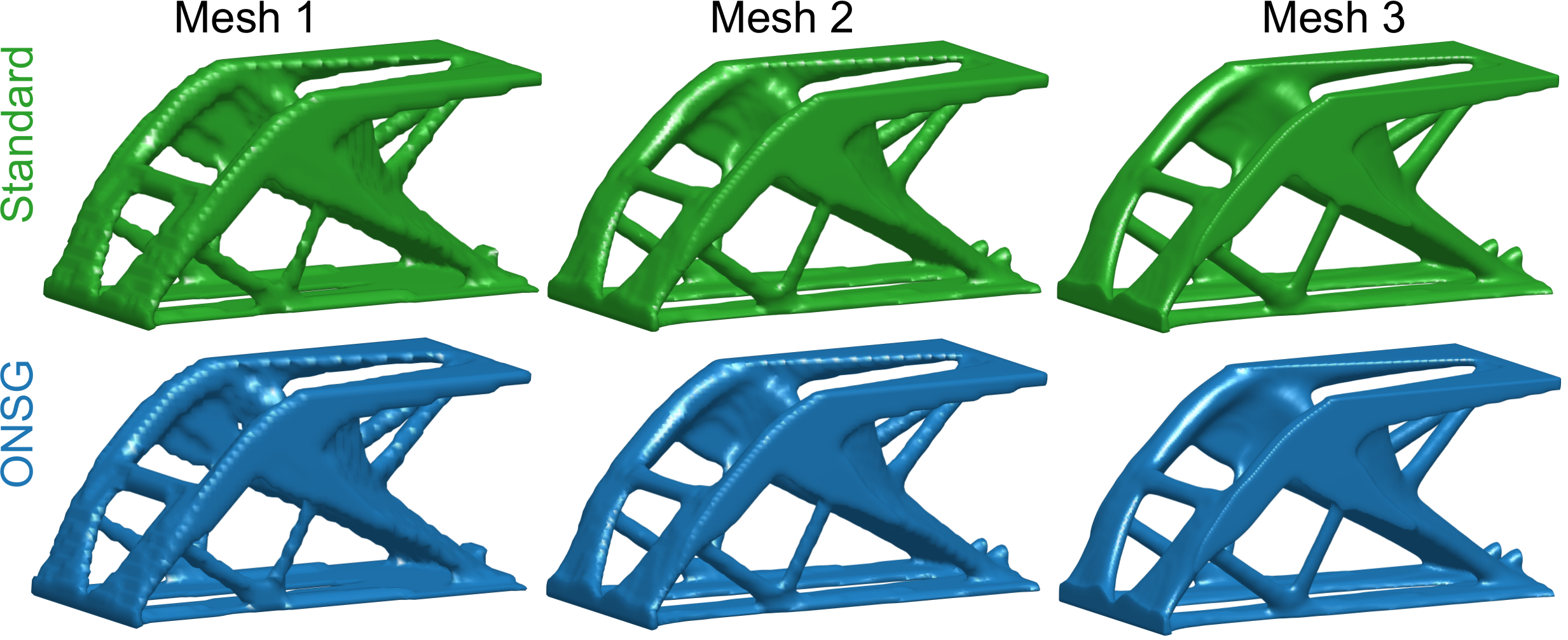}
\caption{Final Cantilever designs obtained by the standard optimizer and our method.}
\label{fig:cantilever_design_res}
\end{figure}

\begin{figure}[t]
\centering
\includegraphics[width=0.48\textwidth]{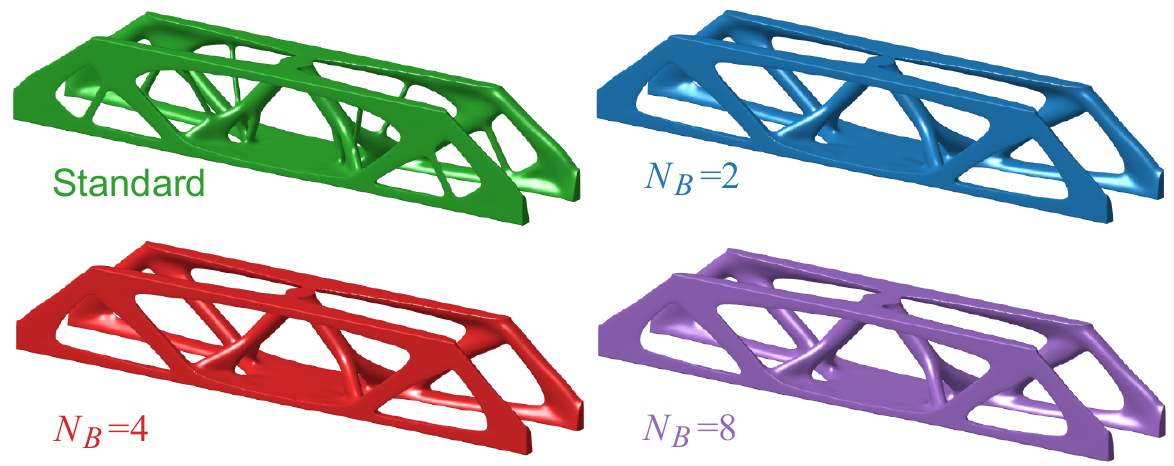}
\caption{Final MBB designs obtained by the standard optimizer and our method.}
\label{fig:MBB_design_res}
\end{figure}

In Figure~\ref{fig:cantilever_design_res} and \ref{fig:MBB_design_res}, we visualize the final designs obtained by the standard optimizer and our method. For the Cantilever design  problem, there is no visual difference observed between our design and the standard one, and both get smoother and higher-quality designs on finer meshes. For the MBB design problem, there is minor difference in our designs compared to the standard one. The standard design is more complex, which has several small members. According to the final objectives in Table~\ref{table:objective}, our method actually finds better designs with lower compliance values.

The experimental results above demonstrate that ONSG has comparable or even better performance compared to the standard optimizer.

\setlength\tabcolsep{3pt}
\begin{table}[t]
\centering
\caption{Final objective achieved by the standard optimizer and our method on two benchmark problems. The objective is the lower the better.}
\label{table:objective}
\begin{tabular}{@{}lcccccc@{}}
\toprule
\multirow{3}{*}{\textbf{}} & \multicolumn{3}{c}{\textbf{Cantilever}} & \multicolumn{3}{c}{\textbf{MBB}} \\ \cmidrule(l){2-7} 
 & Mesh 1 & Mesh 2 & \multicolumn{1}{c|}{Mesh 3} & \multicolumn{3}{c}{Mesh 4} \\ \cmidrule(l){2-7} 
 & \multicolumn{3}{c|}{$N_B=5$} & $N_B=2$ & $N_B=4$ & $N_B=8$ \\ \midrule
Standard & 370.11 & 379.54 & \multicolumn{1}{c|}{383.51} & 221.30 & 221.30 & 221.30 \\
ONSG & 368.61 & 379.37 & \multicolumn{1}{c|}{383.75} & 220.28 & 220.52 & 220.92 \\\midrule
Difference & $-0.40\%$ & $-0.05\%$ & \multicolumn{1}{c|}{$0.06\%$} & $-0.46\%$ & $-0.35\%$ & $-0.17\%$ \\ 
\bottomrule
\end{tabular}
\end{table}

\subsection{Model Scalability} \label{subsec:scalability}

ONSG employs the two-scale optimization scheme to decomposes the global structure into local ones, which is important for solving large-scale optimal design problems. To evaluate the scalability of ONSG, and quantitatively show the impact of the two-scale scheme, we record the number of model parameters and the corresponding GPU memory cost of the Cantilever design problem in Table~\ref{table:gpu_memory}. For the single-scale model, as we increase the scale of mesh from Mesh 1 to Mesh 3, the number of parameters in the deep learning model increases drastically and so does the GPU memory required for training the model. Actually the single-scale model runs out of GPU memory on Mesh 2 and Mesh 3. On the contrary, ONSG with the two-scale scheme maintains a constant number of parameters on all meshes. The GPU memory cost is much less compared to the single-scale model, and has only increased by a modest margin, which is caused by more training data being cached on GPU.

The two-scale scheme requires to solve the coarse-scale PDE constraints with additional computational cost. In fact, this extra cost is almost negligible since the dimensionality of the coarse-scale stiffness matrix $\mathbf{K}_C$ is cubically smaller than the original stiffness matrix $\mathbf{K}$, as listed in Table~\ref{table:stats}. Therefore, with this small extra cost, the two-scale scheme has much better scalability and can easily handle large-scale optimization problems on a single GPU.

\setlength\tabcolsep{11pt}
\begin{table}[t]
\centering
\caption{Number of model parameters and the corresponding GPU memory cost of the Cantilever design problem. A hyphen indicates running out of GPU memory.}
\label{table:gpu_memory}
\begin{tabular}{@{}llccc@{}}
\toprule
Scheme &  & Mesh 1 & Mesh 2 & Mesh 3 \\ \midrule
\multirow{2}{*}{Single-scale} & \# params & 175M & 867M & 3B \\
 & GPU memory & 3.9GB & - & - \\ \midrule
\multirow{2}{*}{Two-scale} & \# params & 3.3M & 3.3M & 3.3M \\
 & GPU memory & 0.7GB & 0.7GB & 0.9GB \\ \bottomrule
\end{tabular}
\end{table}

\begin{figure*}
\centering
    \subfigure[Cantilever Mesh 1]{
        \includegraphics[height=126px]{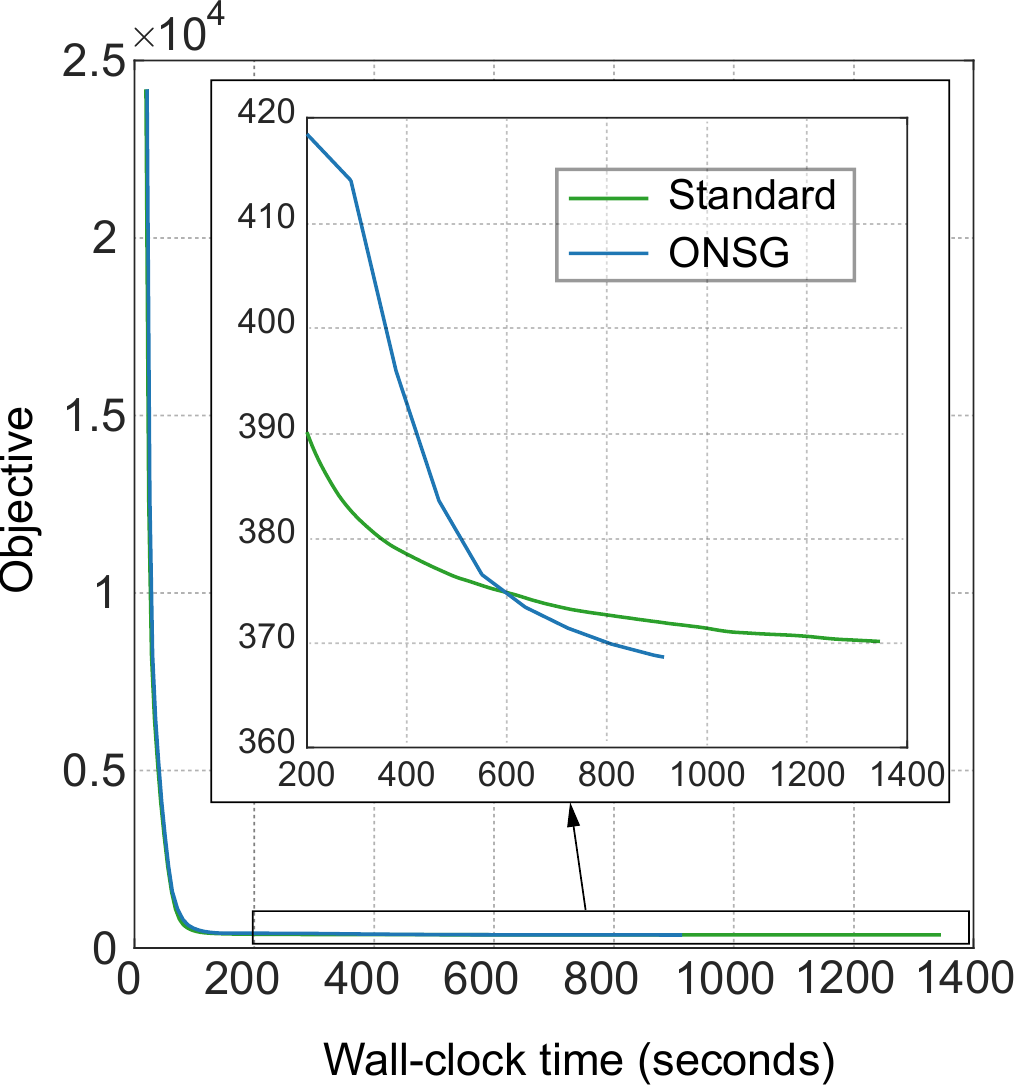}
        \label{fig:cantilever_mesh1_curve}
    }
    \hfill
    \subfigure[Cantilever Mesh 2]{
        \includegraphics[height=126px]{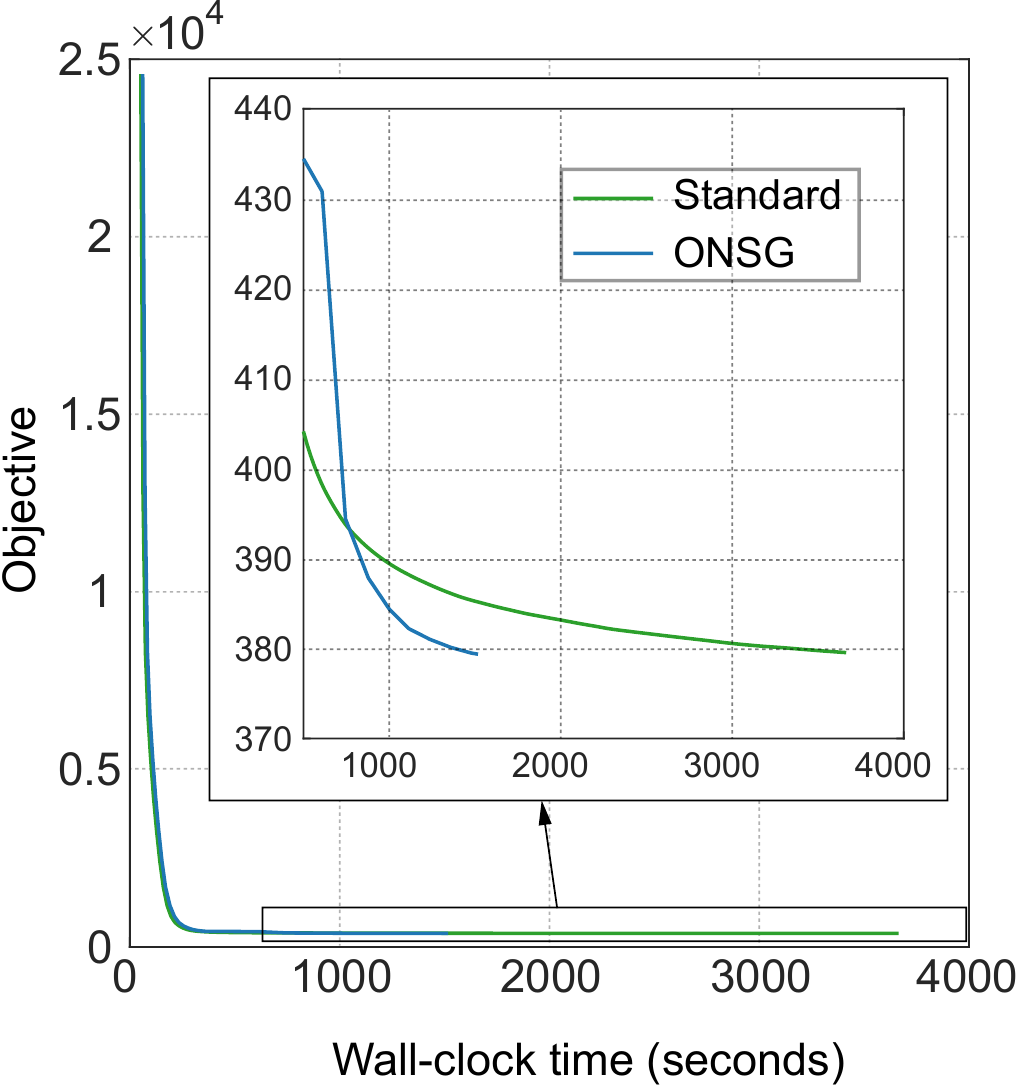}
        \label{fig:cantilever_mesh2_curve}
    }
    \hfill
    \subfigure[Cantilever Mesh 3]{
        \includegraphics[height=126px]{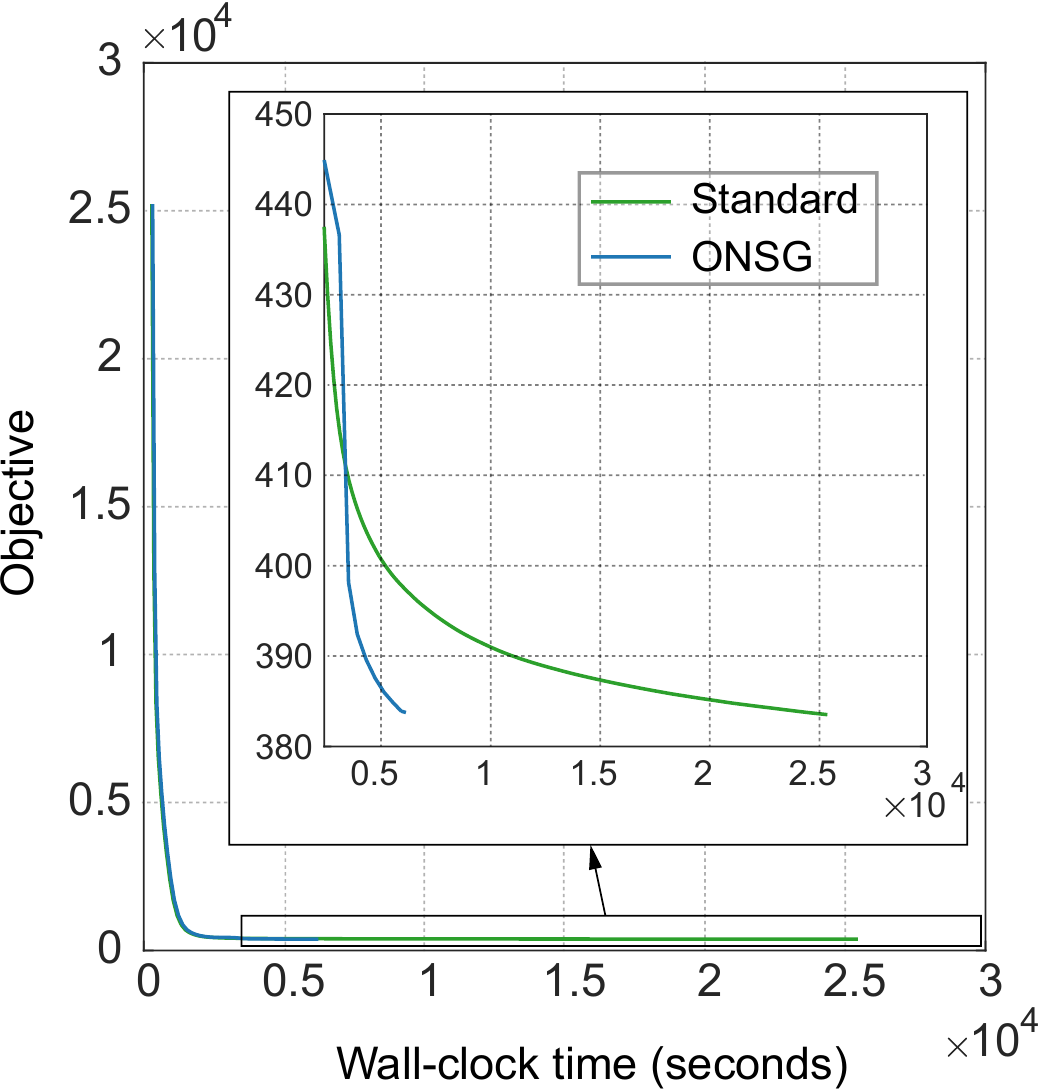}
        \label{fig:cantilever_mesh3_curve}
    }
    \hfill
    \subfigure[MBB Mesh 4]{
        \includegraphics[height=126px]{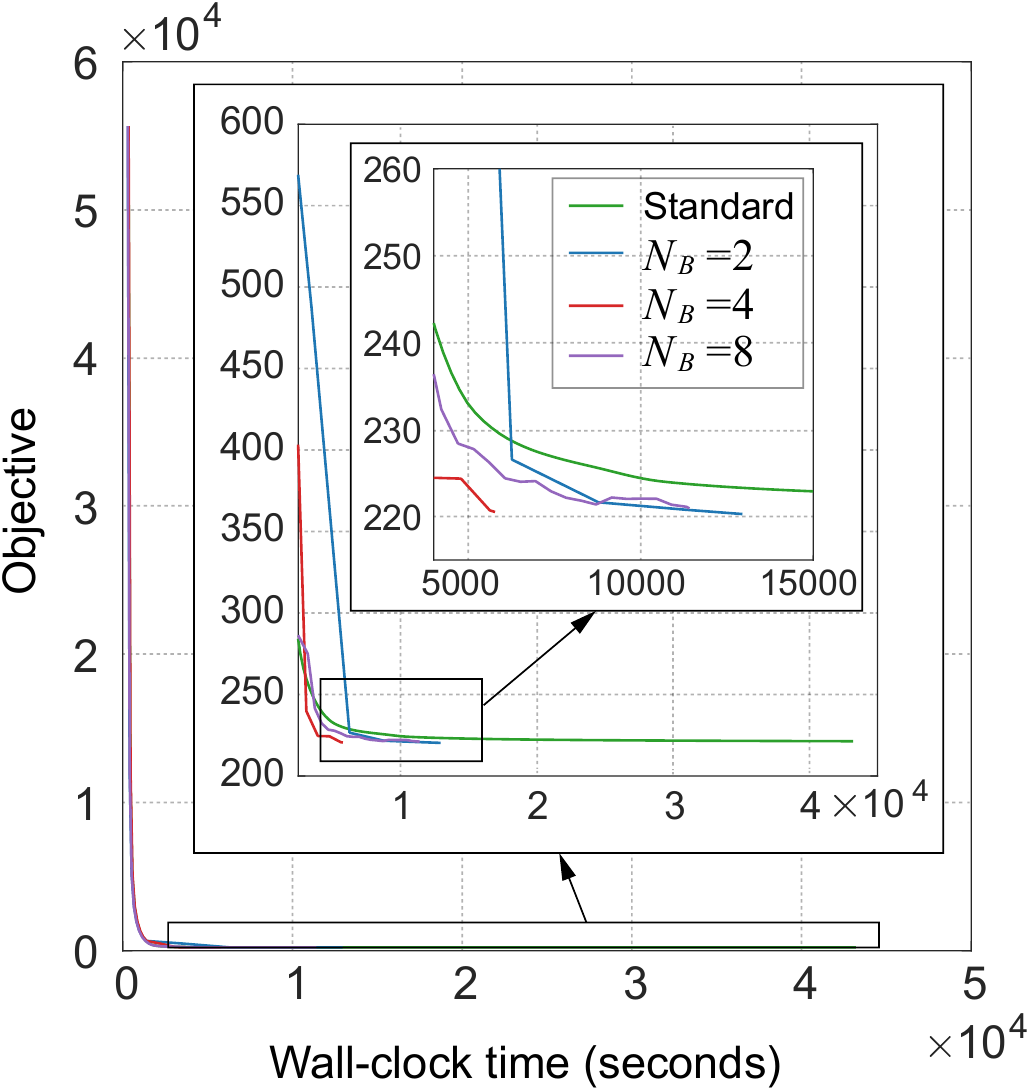}
        \label{fig:MBB_curves}
    }
\caption{Objective curves on the Cantilever and MBB design problems. The objective is the lower the better.}
\label{fig:all_time_curves}
\end{figure*}

\begin{figure}
\centering
	\subfigure[Cantilever]{
	\includegraphics[height=115px]{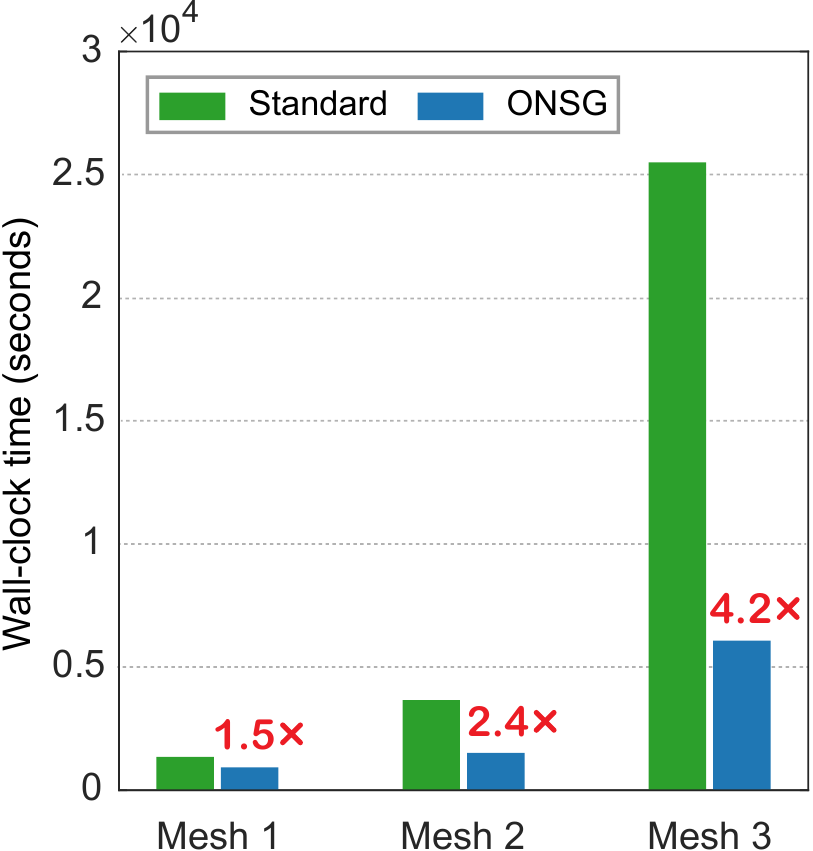}
	\label{fig:cantilever_speedup}
	}
	\hfill
	\subfigure[MBB]{
	\includegraphics[height=115px]{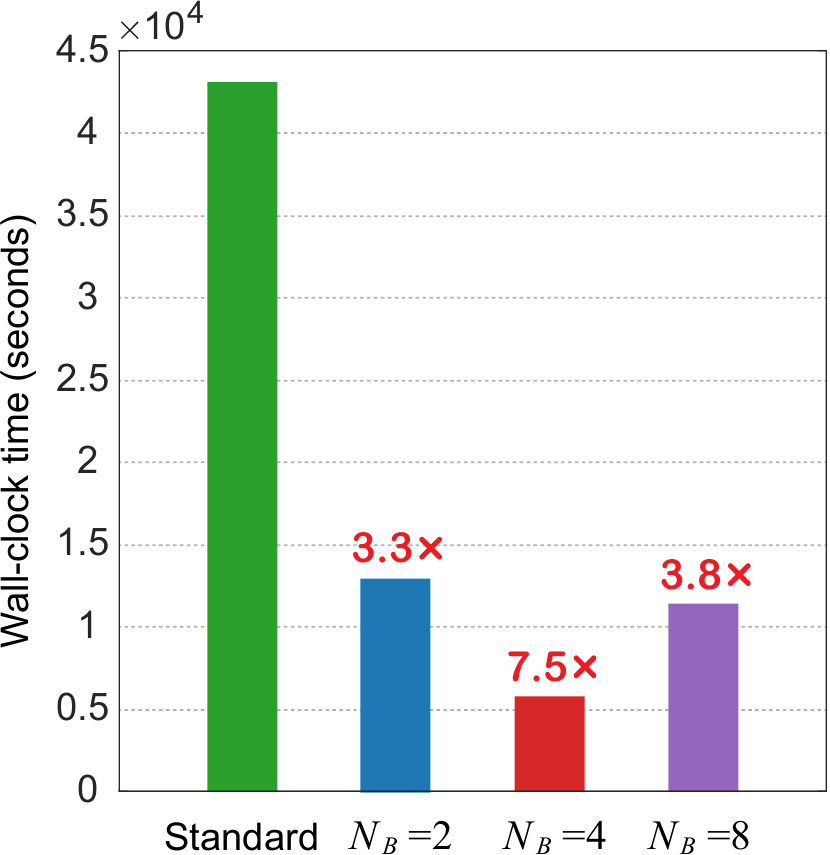}
	\label{fig:MBB_speedup}
	}
\caption{Wall-clock time of the standard optimizer and our method. The corresponding speedup of our method is labeled in red above the bars. For Cantilever, Mesh 1--3 are from the smallest to the largest scale. For MBB, we experiment on a large-scale mesh (Mesh 4), and vary the block size $N_B$.}
\label{fig:speedup}
\end{figure}

\subsection{Model Efficiency}

To evaluate the model efficiency, we time the optimization process for both the standard optimizer and our method, and compare their time cost. We first study the performance on the Cantilever design problem. For Mesh 1--3, the wall-clock time is reported in Figure~\ref{fig:cantilever_speedup}, and the objective curves along with the wall-clock time are plotted in Figure~\ref{fig:cantilever_mesh1_curve}--\ref{fig:cantilever_mesh3_curve}. From these experimental results, we can draw several conclusions: 1) ONSG is significantly faster than the standard optimizer in all cases; 2) ONSG achieves higher speedup on larger scale meshes. On Mesh 3 with over one million design variables, ONSG achieves 4.2x speedup, which is higher than the 2.4x speedup on Mesh 2 and the 1.5x speedup on Mesh 1. For the standard optimizer, when the number of design variables increases, the dimensionality of the stiffness matrix in the PDE constraints $\mathbf{K}$ grows drastically (as listed in Table~\ref{table:stats}), so that on large-scale mesh it is very time-consuming to compute the gradients in every iteration. ONSG employs neural synthetic gradients as a shortcut to skip the expensive computation in most iterations, and it saves more time on larger scale meshes; 3) the objective curves in Figure~\ref{fig:cantilever_mesh1_curve}--\ref{fig:cantilever_mesh3_curve} show that at the beginning ONSG maintains the objective close to the standard one, and then overtakes the standard optimizer and quickly goes down. This is reasonable since the deep learning model has few training data at the beginning, and the synthetic gradients are not that accurate. As the optimization proceeds, more training data is collected to update the model online, so that the synthetic gradients become more accurate and guide the optimization back on track.

\subsection{Further Speedup}
In previous experiments on the Cantilever design problem, we fix the block size $N_B=5$ and our method ONSG achieves up to 4.2x speedup over the standard method. Now we discuss how the block size $N_B$ can have impact on speedup, and explore whether our method can achieve further speedup by varying $N_B$.

In the extreme case, if we set $N_B=1$, there is no difference between the coarse-scale and the original fine-scale PDE constraints, i.e., $\mathbf{K} = \mathbf{K}_C$, which roughly reduces ONSG to the standard optimizer and there should be no speedup anymore. In the other extreme case, if we set $N_B$ to be large enough to cover the entire mesh (suppose we have enough GPU memory), there is only one block to generate the training data. In this case, the learning becomes more difficult due to the high-dimensional input and the small number of training samples. Thus we have to increase the frequency of supervision for the deep learning model, and use larger deep neural networks for model training, which will definitely diminish the speedup of ONSG. Based on our analysis of the two extremes, there should be a trade-off of the block size $N_B$ to achieve the best speedup.

We conduct experiment on the MBB design problem by setting $N_B=2,4,8$ respectively. As reported in Table~\ref{table:objective}, the final objectives achieved are even better than the standard optimizer, showing that ONSG can maintain perfect accuracy with larger block size. The wall-clock time is reported in Figure~\ref{fig:MBB_speedup}. With $N_B=4$, ONSG achieves 7.5x speedup over the standard optimizer, which almost doubles the speedup of other two cases. The corresponding objective curves are plotted in Figure~\ref{fig:MBB_curves}. We can see that the $N_B=2$ curve takes long time to drop, since the coarse-scale PDE constraints are more refined and takes longer to solve. The $N_B=4$ curve is at the bottom, which is the fastest among all the curves. These experimental results demonstrate that a good trade-off of the block size $N_B$ can lead to further speedup of ONSG.

\section{Conclusions and Future Work}

We propose a framework named ONSG to speed up computational morphogenesis. ONSG employs the optimization history to train a deep learning model, and obtains online synthetic gradients from the deep learning model to replace exact gradients as a shortcut to save computational time. To enable the framework to handle large-scale optimization problems, we propose a two-scale optimization scheme, which decomposes the design variables into local subsets but still incorporates global information. With the two-scale scheme, ONSG can efficiently deal with computational morphogenesis problems with millions of design variables. In the future, we plan to formulate other PDE-constrained optimization applications into our framework, such as optimal control and inverse estimation problems.

\bibliography{ref}
\bibliographystyle{IEEEtran}

\end{document}